\documentclass[letterpaper, 10 pt, conference]{ieeeconf}

\IEEEoverridecommandlockouts
\makeatletter
\let\NAT@parse\undefined
\makeatother

    \usepackage{cite}
    \usepackage{amsmath}
    \usepackage{amsthm}
    \usepackage{amsfonts}
    \usepackage{amssymb}
    \usepackage{graphicx}
    \usepackage{array}
    \usepackage[dvipsnames]{xcolor}
    \usepackage{mathrsfs}
    \usepackage{multicol}
    \usepackage{multirow}
    \usepackage{arydshln}
    \usepackage{diagbox}
    \usepackage{mathtools}
    \usepackage{algorithm}
    \usepackage{algpseudocode}
    \algtext*{EndIf}
    \algtext*{EndFor}
    \algtext*{EndWhile}

\usepackage{duckuments}
\usepackage{amsmath}
\usepackage{tabularx}
\usepackage{booktabs}
\usepackage{makecell}
\usepackage{array}

\usepackage{colortbl}
\usepackage{tikz}
\usepackage{array}
\usepackage{graphicx}
\usepackage{multicol}
\usepackage[normalem]{ulem}

\usepackage{subcaption}

\usepackage[bookmarks=true]{hyperref}

        \newcommand{\inv}{^{-1}}
        \newcommand{\abs}[1]{\left|#1\right|}

        \newcommand{\of}{\circ}

        \newcommand{\set}[1]{\left\{#1\right\}}

        \newcommand{\reals}{\mathbb{R}}
        \newcommand{\R}{\reals}

        \newcommand{\norm}[1]{\abs{\abs{#1}}}
        \newcommand{\tpose}{^{T}}
        
        \DeclareMathOperator{\trace}{tr}

        \DeclareMathOperator{\argmin}{argmin}

        \newcommand{\paren}[1]{\left(#1\right)}

        \newcommand{\ptxt}[1]{\textrm{\textnormal{#1}}}
        
        \newcommand{\mc}[1]{\mathcal{#1}}
        \newcommand{\ms}[1]{\mathscr{#1}}
        
        \newtheorem{theorem}{Theorem}

    \usepackage{letltxmacro}
    \LetLtxMacro\orgvdots\vdots
    \LetLtxMacro\orgddots\ddots

    \makeatletter
    \DeclareRobustCommand\vdots{%
        \mathpalette\@vdots{}%
    }
    \newcommand*{\@vdots}[2]{%
        \sbox0{$#1\cdotp\cdotp\cdotp\m@th$}%
        \sbox2{$#1.\m@th$}%
        \vbox{%
            \dimen@=\wd0 %
            \advance\dimen@ -3\ht2 %
            \kern.5\dimen@
            \dimen@=\wd2 %
            \advance\dimen@ -\ht2 %
            \dimen2=\wd0 %
            \advance\dimen2 -\dimen@
            \vbox to \dimen2{%
                \offinterlineskip
                \copy2 \vfill\copy2 \vfill\copy2 %
            }%
        }%
    }
    \DeclareRobustCommand\ddots{%
        \mathinner{%
            \mathpalette\@ddots{}%
            \mkern\thinmuskip
        }%
    }
    \newcommand*{\@ddots}[2]{%
        \sbox0{$#1\cdotp\cdotp\cdotp\m@th$}%
        \sbox2{$#1.\m@th$}%
        \vbox{%
            \dimen@=\wd0 %
            \advance\dimen@ -3\ht2 %
            \kern.5\dimen@
            \dimen@=\wd2 %
            \advance\dimen@ -\ht2 %
            \dimen2=\wd0 %
            \advance\dimen2 -\dimen@
            \vbox to \dimen2{%
                \offinterlineskip
                \hbox{$#1\mathpunct{.}\m@th$}%
                \vfill
                \hbox{$#1\mathpunct{\kern\wd2}\mathpunct{.}\m@th$}%
                \vfill
                \hbox{$#1\mathpunct{\kern\wd2}\mathpunct{\kern\wd2}\mathpunct{.}\m@th$}%
            }%
        }%
    }
    \makeatother

\usepackage{cleveref}
\Crefname{figure}{Fig.}{Figs.}
\Crefname{equation}{Eq.}{Eqs.}
\Crefname{lemma}{Lemma}{Lemmata}
\Crefname{proposition}{Proposition}{Propositions}
\Crefname{assumption}{Assumption}{Assumptions}
\Crefname{theorem}{Theorem}{Theorems}
\Crefname{section}{Section}{Sections}
\Crefname{subsection}{Subsection}{Subsections}
\Crefname{appendix}{Appendix}{Appendices}
\Crefname{corollary}{Corollary}{Corollaries}

\DeclareMathOperator{\SE}{SE}
\newcommand{\monogram}[3]{{}^{#2}\!#1^{#3}}
\DeclareMathOperator{\FK}{FK}
\DeclareMathOperator{\IK}{IK}
\newcommand{\EIK}{\widehat{\IK}}
\newcommand{\pd}[2]{\frac{\partial #1}{\partial #2}}
\newcommand{\tpd}[2]{\tfrac{\partial #1}{\partial #2}}
\newcommand{\colsep}{\hspace{0.5em}}

\DeclareMathOperator{\lb}{lb}
\DeclareMathOperator{\ub}{ub}
\newcommand{\LEFT}{\mathrm{left}}
\newcommand{\RIGHT}{\mathrm{right}}
\newcommand{\base}{\mathrm{base}}
\newcommand{\torso}{\mathrm{torso}}
\newcommand{\full}{\mathrm{full}}
\AtBeginDocument{%
    \setlength{\abovedisplayskip}{5pt plus 2pt minus 2pt}%
    \setlength{\belowdisplayskip}{5pt plus 2pt minus 2pt}%
    \setlength{\abovedisplayshortskip}{2pt plus 2pt}%
    \setlength{\belowdisplayshortskip}{4pt plus 2pt minus 2pt}%
}

\AtBeginDocument{%
    \setlength{\textfloatsep}{8pt plus 2pt minus 2pt}%
    \setlength{\dbltextfloatsep}{8pt plus 2pt minus 2pt}%
    \setlength{\floatsep}{8pt plus 2pt minus 2pt}%
    \setlength{\dblfloatsep}{8pt plus 2pt minus 2pt}%
    \setlength{\intextsep}{8pt plus 2pt minus 2pt}%
    \setlength{\abovecaptionskip}{4pt}%
    \setlength{\belowcaptionskip}{0pt}%
}

\makeatletter
\AtBeginDocument{%
    \def\section{\@startsection{section}{1}{\z@}{1.2ex plus 0.2ex minus 0.5ex}%
        {0.6ex plus 0.1ex minus 0ex}{\normalfont\normalsize\centering\scshape}}%
    \def\subsection{\@startsection{subsection}{2}{\z@}{1.2ex plus 0.2ex minus 0.5ex}%
        {0.6ex plus 0.1ex minus 0ex}{\normalfont\normalsize\itshape}}%
}
\makeatother

\usepackage{color-edits}

\definecolor{darkgreen}{rgb}{0.0, 0.75, 0.0}
\addauthor{tc}{blue}
\addauthor{ss}{darkgreen}
\addauthor{tm}{purple}
\addauthor{nr}{orange}
\addauthor{rt}{red}
\addauthor{hb}{magenta}

\title{\LARGE \bf
Planning along Differentiable Charts of Constraint Manifolds with General-Purpose IK Solvers
}

\author{Thomas Cohn*, Seiji Shaw*, Harel Biggie, Travis Manderson, Nicholas Roy, and Russ Tedrake
\thanks{
* denotes equal contribution.
This project was supported by the National Science Foundation Graduate Research Fellowship Program under Grant No. 2141064, MIT Siegel Family Quest for Intelligence, Natural Sciences and Engineering Research Council of Canada (NSERC), and Army Research Laboratory under Cooperative Agreement Number W911NF-17-2-0181.
Any opinions, findings, and conclusions or recommendations expressed in this material are those of the author(s) and do not necessarily reflect the views of the National Science Foundation or the other sponsors acknowledged in this work.
The authors are with the Computer Science and Artificial Intelligence Laboratory (CSAIL), Massachusetts Institute of Technology, Cambridge, MA, USA. Corresponding author: {\tt\small tcohn@mit.edu}}}

\begin{document}

\maketitle
\thispagestyle{empty}
\pagestyle{empty}

\begin{abstract}
Planning trajectories for robot manipulators under kinematic equality constraints restricts feasible motions to a measure-zero submanifold of the configuration space, requiring special algorithmic treatment.
A promising strategy is parametrizing the set of feasible configurations using analytic inverse kinematics (IK).
Bespoke analytic IK functions can be written to be differentiable, a necessary property for gradient-based trajectory optimization.
But the vast majority of IK functions are computed by automated meta-solvers like IKFast, and are difficult to modify for differentiability.
We present a new approach for computing gradients of analytic IK parameterizations: we leverage the inverse function theorem to recover the desired gradients from the ordinary forward kinematic Jacobian.
Furthermore, we present a least-squares domain extension and an optimization-amenable description of the reachability constraint, which preserves gradient signal outside the reachable workspace.
We demonstrate the efficacy of our approach through numerical experiments and downstream tasks, including a hardware demonstration of an RB-Y1 picking up a box and placing it on a table.
Project website: \href{https://cohnt.github.io/inverse-function-theorem-parameterization/}{https://cohnt.github.io/inverse-function-theorem-parameterization/}

\end{abstract}

\section{Introduction}
\label{sec:introduction}

Motion planning algorithms often inherently model a robot's configuration space (C-space) as a Euclidean space.
But this simple model breaks down for tasks where the robot must adhere to certain equality constraints, including carrying an object with two hands~\cite{krebs2022bimanual}, opening a door~\cite{berenson2011task}, or keeping both feet fixed on the ground~\cite{dai2014whole}.
We specifically focus on equality constraints on end-effector (EE) pose, which are also induced by robot manipulation algorithms that reason about the robot as a floating hand~\cite{khatib1987unified,shridhar2023perceiver,chi2024universal}.

When applying an equality constraint to the end-effector path of a motion plan, the set of constraint-satisfying configurations is reduced to a measure-zero subset of C-space, often referred to as the \emph{constraint manifold}.
This presents a fundamental challenge to motion planning algorithms.
Sampling-based planners can no longer rely on rejection sampling to find paths, instead requiring specialized sampling and interpolation procedures~\cite{kingston2018sampling}.
Trajectory optimizers must handle a challenging nonlinear equality constraint, making them highly dependent on a good initial guess~\cite[\S{}VI.D]{bordalba2022direct}.
Either way, the resulting trajectory often must be post-processed to ensure the constraint is satisfied to a high tolerance, or used with a compliant control algorithm~\cite{bonilla2017noninteracting}.

\begin{figure}
    \centering
    \includegraphics[width=\linewidth]{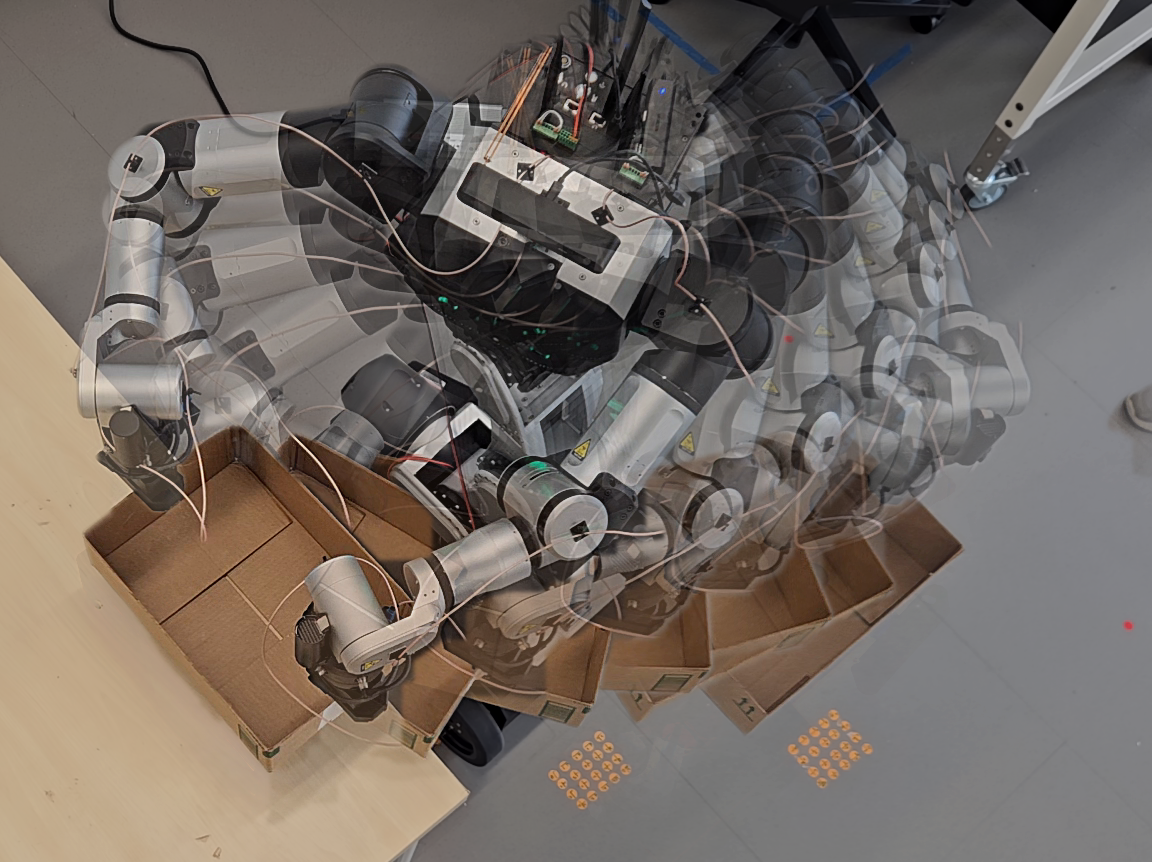}
    \caption{
        A constraint-satisfying motion planned with our minimal coordinates trajectory optimization framework, using the RB-Y1 bimanual manipulator to pick up a box from the floor and place it on a table.
    }
    \label{fig:teaser}
\end{figure}

Parameterization approaches simplify the problem by eliminating the equality constraint, yielding a positive measure feasible set bounded by inequality constraints.
This enables the direct application of ordinary planning algorithms without modification.
A recent line of work examined using \emph{analytic inverse kinematics} (IK) to directly parameterize the constraint manifold.
Analytic IK serves a principled solution to the one-to-many property of IK, taking in additional arguments to return a unique solution for an EE target.
Together with automatic differentiation (autodiff) through the IK mapping, a trajectory optimizer can represent the robot's path in the parameterized space while imposing costs and constraints in configuration space.
This approach has been effective for collision-free constrained planning~\cite{cohn2024constrained} and as a tool for solving complex IK problems like humanoid stability~\cite{cohn2026framework}.
Some commonly-used robot arms have bespoke, handwritten analytic IK solutions~\cite{faria2018iiwa,he2021frankapanda}, but otherwise, the standard way forward is to use automated meta-solvers that parse the kinematic equations and generate a solution~\cite{diankov2010ikfast,srinivasa2026ssik}.
But these solvers do not produce autodiff-compatible solutions, \emph{leaving most robots (including the RB-Y1 shown in \Cref{fig:teaser}) without the gradients they need, and therefore incompatible with the parameterized optimization framework.}

Rather than editing the IK solution, an onerous task given the optimized implementations produced by automated IK solvers, we present an approach for differentiating through any black-box IK function.
We construct the augmented forward kinematics (FK)~\cite{baillieul1985kinematic,elias2024redundancy}, a variant of ordinary FK that also returns the self-motion parameters corresponding to an analytic IK formulation. 
The augmented FK Jacobian has a simple expression, and the inverse function theorem (IFT) can be used to calculate the IK Jacobian.

Even with this ability to compute the Jacobians of black-box IK solvers, the domain of an analytic IK function can be limited, so gradients are not well-defined for end-effector poses outside the reachable workspace.
Nonlinear trajectory optimizers do not guarantee workspace reachability at each iteration of the optimization~\cite{wachter2006ipopt,gill2005snopt}, leading to a catastrophic error.
Some IK solutions return approximate solutions outside the reachable workspace~\cite{elias2025ik,ostermeier2025automatic}, but popular automatic IK solvers return ``no solution'' for non-reachable targets~\cite{diankov2010ikfast,srinivasa2026ssik}.
This precludes their usage with many optimization frameworks, even if gradients are available within the reachable workspace from the IFT.

We extend the domain of an analytic IK function by leveraging least-squares solutions (or efficient approximations), obtaining a closed-form Jacobian for end-effector poses outside the reachable workspace.
This preserves the gradient flow outside the reachable workspace, allowing the optimizer to restore feasibility.
Inspired by the numerical IK literature, we present efficient approximations to this Jacobian.
Finally, we describe an optimization-amenable description of the domain of an IK function, which is active only on the boundary of the feasible set.
Empirically, we demonstrate that the overall framework begins to close the gap between bespoke and automatically generated analytic IK solutions through numerical studies, downstream planning tasks, and experiments with generic IK solvers.
We further demonstrate the method's practical applicability in 20 hardware pick-and-place trials on an RB-Y1 robot, moving a box from various locations on the floor to an adjacent table (\Cref{fig:teaser}).

\section{Related Work}
\label{sec:related_work}

The principles of motion planning with holonomic equality constraints are well-established.
These constraints naturally describe the constraint manifold in implicit form as the zero-level set of some function.
This representation is less amenable to planning than an explicit parameterization, which is notoriously difficult to construct ~\cite[p.~168]{lavalle2006planning}, \cite[p.~151]{siciliano2008springer}, \cite[p.~28]{lynch2017modern}.
Thus, much of the existing literature has focused on utilizing the implicit representation.

Sampling-based planning, one of the two main motion planning paradigms, has been extensively applied to the kinematically-constrained case~\cite{kingston2018sampling,kingston2019exploring}.
Approaches have largely relied on specialized procedures to draw samples from the manifold, such as generic nonlinear optimization or inverse kinematics~\cite{yakey2001randomized,berenson2011task,iyer2026vectorizing}.
Alternatively, continuation approaches for building a piecewise linear approximation can be interleaved with the sampling~\cite{jaillet2012path}.
Mature frameworks for sampling-based planning with constraints include OMPL~\cite{sucan2012open}, CuikSuite~\cite{porta2014cuik}, and HPP~\cite{mirabel2016hpp}.

On the other hand, many trajectory optimizers can directly handle arbitrary constraints, although their performance is generally dependent on the initial guess~\cite[\S{}VI.D]{bordalba2022direct}.
The theory of trajectory optimization on manifolds is well-understood~\cite{bonalli2019trajectory}, and there are principled methods for planning on special manifolds~\cite{teng2025riemannian}.
But the current best methods for optimization-based planning on generic implicitly-defined manifolds combine direct collocation with the local approximations (projection or piecewise-linear approximation) used by sampling-based planning~\cite{bordalba2022direct}.

\section{Background}
\label{sec:background}

We use monogram notation~\cite[\S 3.1]{russtedrake2024manipulation} to describe poses: $\monogram{X}{A}{B}\in\SE(3)$ is the pose of frame $B$ relative to (and expressed in) frame $A$.
When this pose is a function of a variable $q$, we write $\monogram{X}{A}{B}(q)$.
The forward kinematics are $\FK:\R^n\to\SE(3)$.
An analytic IK function is a mapping
\begin{equation}
    \label{eq:analytic_ik}
    \IK:\mc U\times\Psi\times\ms K\to\R^n,
\end{equation}
where $\mc U\subseteq\SE(3)$ is a set of reachable end-effector poses, $\Psi$ describes the continuous self-motions and $\ms K$ enumerates the discrete self-motions (e.g. ``elbow up'' vs ``elbow down'')~\cite{burdick1989inverse}.
In practice, we restrict ourselves to some $\kappa\in\ms K$ and drop the discrete self-motion from the notation~\cite{cohn2024constrained}.

Analytic IK allows us to chart the constraint manifold resulting from certain closed kinematic linkages: if removing analytically-solvable subchains from the linkage graph results in a collection of disconnected kinematic trees, we naturally have a parameterization~\cite{han2001akinematics}.
Autodiff through this mapping enables gradient-based optimization with variables in the minimal coordinates, but with costs and constraints in C-space.
For example, the decision variables for our grasp selection experiment (\Cref{sec:experiments:generic:eaik}) are the end-effector pose and self-motion parameter, but we impose joint limits and a joint-centering objective in C-space.

Robot manipulators often have limited workspaces, so optimization problems in the parameterized coordinates must impose a \emph{reachability constraint}.
One way to enforce reachability is with \emph{direct reachability} constraints, which require the achieved end-effector transform matches the desired~\cite{cohn2024constrained}:
\begin{equation}
    \norm{\FK(\IK(\monogram{X}{W}{G},\psi))-\monogram{X}{W}{G}}_F^2\le 0,
    \label{eq:direct_reachability}
\end{equation}
but this constraint is active for all feasible configurations, which can be detrimental for many optimizers.
When privileged information about the internals of the analytic IK solution is available, we can construct a better formulation of reachability (so-called \emph{probing functions}~\cite[\S{}IV]{cohn2026framework}).
Unfortunately, this approach is impractical for IK solutions produced by automated meta-solvers like IKFast.

For a constraint manifold defined implicitly as $\mc M=\set{q\in\R^n:F(q)=0}$, a minimum-cost path (with respect to a cost functional $L$) between $q_0,q_1\in\mc M$ can be found via
\begin{equation}
    \label{eq:trajopt_extrinsic}
    \renewcommand{\arraystretch}{0.95}
    \begin{array}{rlr}
        \argmin & L(\gamma) \\
        \ptxt{s.t.} & \gamma\in\mc C^1([0,1],\R^n), \\
        & \gamma(s)\in\mc M, & \forall s\in[0,1], \\
        & \gamma(s)\ptxt{ is collision free}, & \forall s\in[0,1], \\
        & \gamma(0)=q_0,\gamma(1)=q_1.
    \end{array}
\end{equation}
Given a parameterization $\phi:\mc V\to\mc M$ with $\mc V\subseteq\R^m$, $m\le n$, and points $\tilde q_0,\tilde q_1$ such that $\phi(\tilde q_0)=q_0$ and $\phi(\tilde q_1)=q_1$, this problem can be rewritten as
\begin{equation}
    \label{eq:trajopt_intrinsic}
    \renewcommand{\arraystretch}{0.95}
    \begin{array}{rlr}
        \argmin & L(\phi\of\tilde\gamma) \\
        \ptxt{s.t.} & \tilde\gamma\in\mc C^1([0,1],\mc V), \\
        & (\phi\of\tilde \gamma)(s)\ptxt{ is collision free}, & \forall s\in[0,1], \\
        & \tilde\gamma(0)=\tilde q_0,\tilde\gamma(1)=\tilde q_1.
    \end{array}
\end{equation}
For many constrained trajectory optimization problems, the equality constraint $\gamma(s)\in\mc M$ is the ``hardest'' constraint, so its elimination via an IK parameterization results in an easier-to-solve problem.
We can interpret an optimization IK problem in the same way, by optimizing for a single point instead of a trajectory.

\section{Methodology}
\label{sec:methodology}

Suppose we are given an analytic IK implementation that exposes the discrete and redundant self-motion but does not support autodiff.
A lack of explicit differentiability would prevent the use of gradient-based optimization, as we need the Jacobian of this mapping (or some other way of computing Jacobian-vector products) to obtain the gradient of the costs and constraints.

\subsection{Applying the Inverse Function Theorem}
\label{sec:methodology:ift}

We can leverage the inverse function theorem (IFT)~\cite[Thm.~2.11]{spivak2018calculus} to compute the Jacobian of $\IK$ in terms of the derivative of $\FK$, which can easily be computed with standard robotics toolboxes.

\begin{theorem}[Inverse Function Theorem]
If $f:\R^n\to\R^n$ is continuously differentiable at a point $p$ and has a full-rank Jacobian, then there is an inverse function $f\inv$ defined on a neighborhood of $f(p)$, and $[Df(p)]\inv=D(f\inv)f(p)$.
\end{theorem}

For a nonredundant manipulator, the EE pose uniquely identifies the joint angles up to the choice of discrete self-motion $\kappa\in\ms{K}$, so $\IK\inv=\FK$.
For a redundant manipulator, we call the inverse of the analytic IK function the \emph{augmented} forward kinematics (inspired by the language of Elias and Wen~\cite{elias2024redundancy}).
We write $\FK_A:\R^n\to\SE(3)\times\Psi\times\ms K$ and restrict to a single discrete self-motion $\kappa\in\ms K$; its Jacobian will have the block structure
\begin{equation}
    \label{eq:augmented_jacobian_block_structure}
    D\FK_A(q)=\begin{bmatrix}
        D\FK(q)\\
        D\FK_\psi(q)
    \end{bmatrix},
\end{equation}
where $\FK_\psi$ maps the joint angles $q$ to their corresponding self-motion parameter.
If joint angles are used to parameterize the self motion (as in IKFast~\cite{diankov2010ikfast}), the rows of $D\FK_\psi(q)$ are standard basis vectors.
For a 7DoF arm using the shoulder-elbow-wrist parameterization, it is still simple to compute this Jacobian~\cite{elias2024redundancy}.

Now, we leverage $D \FK_A(q)$ to compute gradients of the analytic IK mapping.
Suppose we have decision variables $(X,\psi)$ and partials $(\pd{X}{y},\pd{\psi}{y})$ for some other variable $y$.
We obtain $q=\IK(X,\psi)$, and want to compute $\pd{q}{y}=D\IK(X,\psi)(\pd{X}{y},\pd{\psi}{y})\tpose$.
Due to the IFT, $D\FK_A(q)=[D\IK(X, \psi)]\inv$, so we find $\pd{q}{y}$ by solving the linear system
\begin{equation}
    \label{eqn:ift_linear_sys}
    D\FK_A(q)\tpd{q}{y}=\paren{\tpd{X}{y},\tpd{\psi}{y}}\tpose.
\end{equation}
This is analogous to how differential inverse kinematics (DiffIK) resolves kinematic redundancy locally, but our use of the augmented Jacobian ensures global consistency.
For non-reachable configurations (and representational singularities~\cite{elias2024redundancy}), $D\FK_A$ loses rank, and the IFT breaks down.
In this case, we fall back to an approximate solution (details deferred to \Cref{sec:methodology:domain_extension}).

\textbf{A worked example.} We now discuss how to apply the IFT machinery to a parameterization for constrained bimanual motion planning~\cite{cohn2024constrained}.
Let $q_c,q_s\in\R^7$ be the joint angles of the ``controlled'' and ``subordinate'' arms, $E_c$ and $E_s$ their end-effector frames, and let $\psi_s\in\R$ be the subordinate arm's continuous self-motion parameter.
Let $\FK_c$ be the forward kinematics of the controlled arm and $\FK_A$ refer to the augmented forward kinematics of the subordinate arm.
We compute the desired pose of the subordinate arm's end-effector in terms of $q_c$ via $\monogram{X}{W}{E_s}(q_c)=\monogram{X}{W}{E_c}(q_c)\monogram{X}{E_c}{E_s}$, where $\monogram{X}{E_c}{E_s}$ is the fixed transform between the end-effectors.
Then we compute $q_s=\IK(\monogram{X}{W}{E_s}(q_c),\psi_s)$, obtaining the full configuration $(q_c,q_s)=\phi(q_c,\psi_s)$.
Thus, $\phi$ parameterizes the constraint manifold, and we wish to compute the derivatives
\begin{equation*}
    \tpd{(q_c, q_s)}{y} = D\phi(q_c,\psi_s)\paren{\tpd{q_c}{y}, \tpd{\psi}{y}}\tpose.
\end{equation*}

First, the Jacobian of $\phi$ has the organized block structure
\begin{align*}
    \hspace{-0.5em}
    D\phi
    &=
    \pd{(q_c,q_s)}{(q_c,\psi_s)}
    =
    \left[
    \begin{array}{c}
        \pd{q_c}{q_c} \quad \pd{q_c}{\psi_s}
        \vphantom{\tfrac{}{)}}
        \\ \hdashline
        \vphantom{\tfrac{)}{}}
        \pd{q_s}{(q_c,\psi_s)}
    \end{array}
    \right]\\
    &=
    \left[
    \begin{array}{c}
        \displaystyle I_{7\times 7} \quad 0_{7\times 1}
        \\ \hdashline
        \pd{q_s \strut}{(\monogram{X}{W}{E_s},\psi_s) \strut}
        \pd{(\monogram{X}{W}{E_s},\psi_s) \strut}{(q_c,\psi_s) \strut},
    \end{array}
    \right]
    \\
    &=
    \left[
    \begin{array}{c}
        I_{7\times 7} \qquad\qquad\qquad 0_{7\times 1}
        \\[0.3em] \hdashline \noalign{\vspace{0.3em}}
        \displaystyle(D\FK_A(q_s))\inv
        \begin{bsmallmatrix}
            \monogram{X}{E_c}{E_s}D\FK_c(q_c) & 0_{6\times 1}\\
            0_{1\times 7} & 1_{1\times 1}
        \end{bsmallmatrix}
    \end{array}
    \right],
\end{align*}
expressed in known or easily-computable quantities.

Next, we compute the Jacobian-vector product $\pd{(q_c,q_s)}{y}=D\phi\pd{(q_c,\psi_s)}{y}$.
We could compute $D \phi$ directly by inverting $D\FK_A(q_s)$, but symbolic manipulation reveals a more efficient strategy: finding $\frac{\partial q_s}{\partial y}$ by solving the linear system
\begin{align*}
    \hspace{-0.5em}
    D\FK_A(q_s)\pd{q_s}{y}
    & =
    \begin{bmatrix}
        \monogram{X}{E_c}{E_s}D\FK_c(q_c) & 0_{6\times 1}\\
        0_{1\times 7} & 1_{1\times 1}
    \end{bmatrix}
    \begin{bmatrix}
        \pd{q_c}{y}
        \vphantom{\tfrac{}{)}}
        \\
        \vphantom{\tfrac{)}{}}
        \pd{\psi_s}{y}
    \end{bmatrix}.
\end{align*}

\subsection{Extending the Domain of IK}
\label{sec:methodology:domain_extension}

Solutions derived from meta-solvers like IKFast return ``no solution'' for non-reachable targets.
\emph{This is catastrophic for general nonlinear optimizers.}
Even if the IK function returns approximate solutions outside the reachable workspace, there are two issues with the IFT strategy: the contribution of the approximation procedure to the gradient is not accounted for, and the boundary of the reachable workspace is composed of singular points, so the IFT itself does not apply.

To examine how to compute the gradients in these conditions, we interpret the various methods for extending the domain of $\IK$ as approximations of
\begin{equation}
    \EIK(\monogram{X}{W}{G},\psi)=\operatorname{argmin}_q\norm{\FK_A(q)-(\monogram{X}{W}{G},\psi)}^2_2.
    \label{eq:extended_ik}
\end{equation}
For reachable $\monogram{X}{W}{G}$, $\EIK(\monogram{X}{W}{G},\psi)=\IK(\monogram{X}{W}{G},\psi)$.
Otherwise, $\EIK$ returns a configuration that achieves the closest end-effector pose in a least-squares sense.

We can compute $D\EIK$ via sensitivity analysis.
Let $q^*=\EIK(\monogram{X}{W}{G},\psi)$ be the optimum, $J_A=D\FK_A(q^*)$ the augmented Jacobian, $r_i=\FK_A(q^*)-(\monogram{X}{W}{G},\psi)$ the augmented task-space residual, and $H_i=D_q[J_A]_i$ the Hessian slice corresponding to the $i$th row of $J_A$.
Then
\begin{equation}
    D\EIK(\monogram{X}{W}{G},\psi)=\Big(J_A\tpose J_A+\sum_{i=1}^nr_iH_i\Big)\inv J_A\tpose.
    \label{eq:extended_ik_gradient}
\end{equation}
Observe that if $\monogram{X}{W}{G}$ is reachable, the residual $r_i=0$, so $D\EIK(\monogram{X}{W}{G},\psi)=\paren{J_A\tpose J_A}\inv J_A\tpose=J_A^\dagger$.
Thus, if $J_A$ is full rank, we recover the IFT formulation.
Also, this matrix inversion (and those in the following approximations) can be avoided with linear system solves, analogous to \eqref{eqn:ift_linear_sys}.

As for the non-reachable case, a nonzero residual means the Hessian term is not eliminated, potentially requiring expensive computations.
However, our experimental results will demonstrate that an approximation of \eqref{eq:extended_ik_gradient} matches or beats its performance.
These approximations are inspired by the rich numerical inverse kinematics literature~\cite{buss2004introduction}, which considers formulations very similar to \eqref{eq:extended_ik_gradient}.
The approximations under consideration for the non-reachable case, in order of increasing complexity, are:

\subsubsection{Zero Gradients}
This effectively matches how \cite{cohn2024constrained} handled derivatives for non-reachable end-effector targets.
\begin{equation}
    \label{eq:gradient_approximation:zero}
    D\EIK(\monogram{X}{W}{G},\psi)\approx 0.
\end{equation}

\subsubsection{Pseudoinverse}
If $J_A$ is singular, this should still preserve gradient flow in the non-singular directions while avoiding a numerical blowup.
\begin{equation}
    \label{eq:gradient_approximation:pseudoinverse}
    D\EIK(\monogram{X}{W}{G},\psi)\approx J_A^\dagger.
\end{equation}

\subsubsection{Levenberg-Marquardt (LM)}
The classic damped least-squares~\cite{wampler1986manipulator}.
For $\lambda$, we consider both \emph{constant} and \emph{singular value thresholding} (SVT), where
$\lambda=\lambda_{\max} ( 1 - ( \frac{\sigma_{\min}(J_A)}{\epsilon} )^2)$ if $\sigma_{\min}(J_A) \le \epsilon$, and $\lambda=0$ otherwise, to only apply damping if $J_A$ is ill-conditioned.
\begin{equation}
    \label{eq:gradient_approximation:lm}
    D\EIK(\monogram{X}{W}{G},\psi)\approx \paren{J_A\tpose J_A + \lambda I}\inv J_A\tpose.
\end{equation}

\subsubsection{Residual Damping}
For a larger residual, the Jacobian at the projected configuration is less informative.
\begin{equation}
    \label{eq:gradient_approximation:residual_damping}
    D\EIK(\monogram{X}{W}{G},\psi)\approx (J_A\tpose J_A + \lambda\norm{r}^2I)\inv J_A\tpose.
\end{equation}

\subsubsection{Anisotropic Damping}
A damping factor is introduced along each singular direction that scales with the magnitude of the residual in that direction.
Let $J_A=U\Sigma V\tpose$ be the SVD, $\Lambda_j=\lambda(\norm{r}^2+3(u_j\tpose r)^2)$, $\Lambda=\operatorname{diag}(\Lambda_j)$. Then
\begin{equation}
    \label{eq:gradient_approximation:anisotropic_damping}
    D\EIK(\monogram{X}{W}{G},\psi)\approx \paren{J_A\tpose J_A+V\Lambda V\tpose}\inv J_A\tpose.
\end{equation}

\subsubsection{Full Newton}
Newton's method is well-understood for least-squares problems~\cite[\S10]{nocedal2006numerical}.
We only use the first $6$ components of the augmented kinematic Hessian, corresponding to the end-effector pose, as these can be obtained in closed-form~\cite{hourtash2005kinematic}.
When self-motion is parameterized by the angle of one or more joints, that Hessian component vanishes; otherwise we would require second-order autodiff.
\begin{equation}
    \label{eq:gradient_approximation:full_newton}
    D\EIK(\monogram{X}{W}{G},\psi)\approx \Big(J_A\tpose J_A+\sum_{i=1}^6r_iH_i+\lambda I\Big)\inv J_A\tpose.
\end{equation}

We emphasize that we do not compute the least-squares solution in practice, but instead rely on efficient approximations: $\arccos$ clipping~\cite{cohn2024constrained}, geometric subproblem least-squares~\cite{elias2025ik,ostermeier2025automatic}, or even a simple bisection search to a canonical reachable target (used for our RB-Y1 experiment).

\subsection{New Reachability Constraints}
\label{sec:methodology:reachability}

The IFT strategy for differentiating through the IK function provides the gradients we need, and the least-squares domain extension gives reasonable solutions (and gradients) even when the current iterate is not yet feasible.
But now the IK solution could be incorrect, so we must prevent this with an additional constraint.
Direct reachability \eqref{eq:direct_reachability} is active everywhere, even on the (relative) interior of the feasible set, so we propose a new reachability constraint that is only active on the boundary.

The boundary of the reachable workspace is composed of critical values, i.e., a configuration in the preimage of the boundary must have low-rank Jacobian~\cite[\S{}3.3]{siciliano2008springer}.
$\EIK$ projects onto the boundary of the reachable set, a property maintained by all three approximations used in our experiments.
Thus, enforcing non-singularity of the kinematic Jacobian is sufficient for enforcing reachability.
Let $q^*=\EIK(\monogram{X}{W}{G},\psi)$ and $J=D\FK(q^*)$ (the ordinary kinematic Jacobian), then our boundary reachability constraint is
\begin{equation}
    b:=-\log\det(JJ\tpose+\epsilon I)\le\tau,
    \label{eq:boundary_reachability}
\end{equation}
for hyperparameters $\epsilon$ and $\tau$, which must be chosen on a per-problem basis.
We use this log-determinant formulation for its well-behaved gradients: the constraint gradient $\pd{b}{(\monogram{X}{W}{G},\psi)}$ can be compiled from the following partial derivatives
\begin{equation}
    \pd{b}{q^*_i}=-2\trace\Big(J\tpose\big(JJ\tpose +\epsilon I\big)\inv\pd{J}{q_i^*}\Big),
    \label{eq:boundary_reachability_gradient}
\end{equation}
together with the chain rule and $D\EIK$ obtained from any of the methods in \Cref{sec:methodology:domain_extension}.
Computing \eqref{eq:boundary_reachability_gradient} requires the kinematic Hessian, a more expensive computation than direct reachability, which only needs the kinematic Jacobian.
But in practice, boundary reachability outperforms direct reachability, thanks in part to its better optimization properties.

\section{Experiments}
\label{sec:experiments}

\begin{figure}[!t]
    \centering
    \begin{subfigure}{0.7\linewidth}
        \centering
        \includegraphics[width=\linewidth]{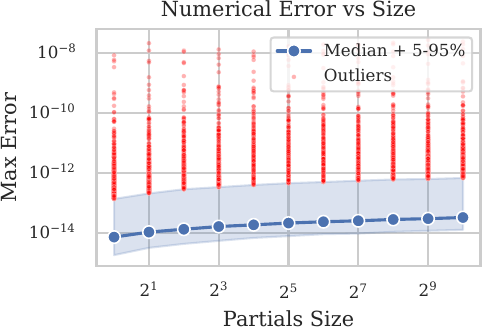}
    \end{subfigure}\\[0.5\baselineskip]
    \begin{subfigure}{0.7\linewidth}
        \centering
        \includegraphics[width=\linewidth]{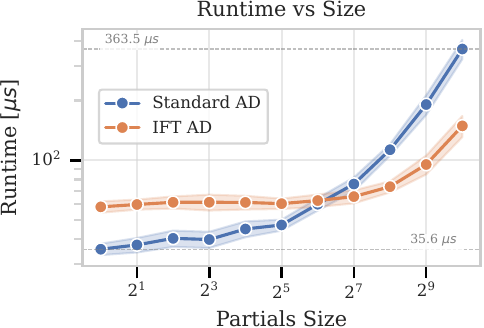}
    \end{subfigure}
    \caption{
        Error (top) and runtime (bottom) of the IFT gradients, in comparison to autodiff, tested on 10\,000 randomly-sampled reachable configurations.
        Axes are log-scale, shading represents the 5-95\% range, and outliers with higher error are indicated in red.
    }
    \label{fig:autodiff_results}
\end{figure}

\begin{figure}[!t]
    \centering
    \includegraphics[width=\linewidth]{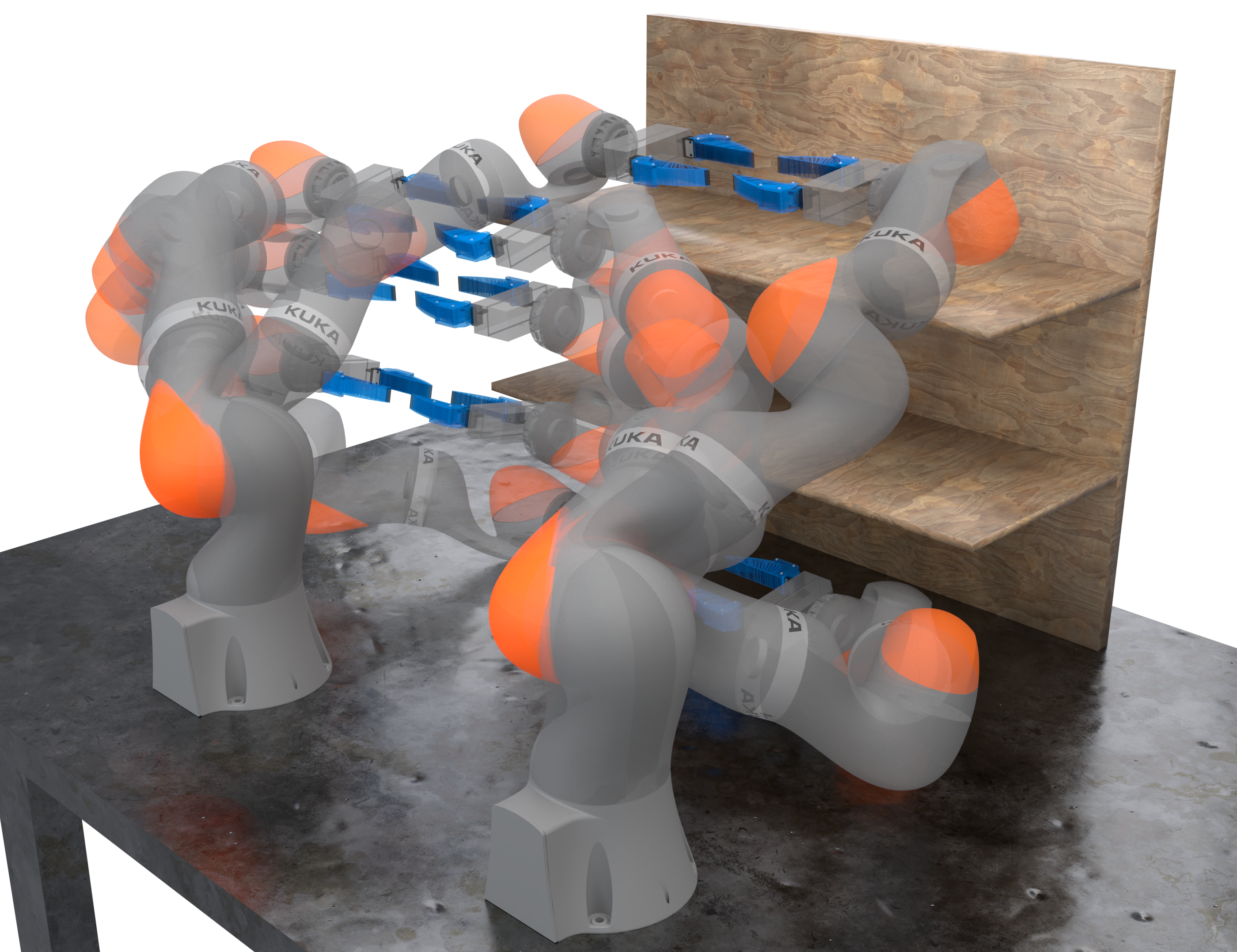}
    \caption{
        A constraint-satisfying motion for a bimanual KUKA iiwa setup~\cite{cohn2024constrained}, planned using trajectory optimization with our proposed IFT gradients.
        Throughout the motion, the EEs must maintain a constant relative transform.
    }
    \label{fig:iiwa_swept_volume}
\end{figure}

\begin{table}[!t]
    \centering
    \caption{
        Downstream runtimes (seconds) by gradient strategy.
    }
    \label{tab:downstream_timing}
    \begin{tabular*}{\columnwidth}{@{}l@{\extracolsep{\fill}}ccc@{}}
    \toprule
    \textbf{Gradient Method} & \textbf{IrisNp2} & \textbf{Trajopt} & \textbf{TOPPRA} \\
    \midrule
    \multicolumn{4}{@{}l}{\textit{Autodiff Baselines}} \\
    Autodiff, Direct Reachability                                                   & 1.14 & 1.54 & 0.59 \\
    Autodiff, Probing Reachability                                                  & 1.03 & 0.98 & 0.60 \\
    Autodiff, Boundary Reachability                                                 & 1.08 & 1.24 & 0.58 \\
    \midrule
    \multicolumn{4}{@{}l}{\textit{Proposed: IFT, Direct Reachability}} \\
    IFT, Zero Gradients \eqref{eq:gradient_approximation:zero}                      & 4.11 & 1.35 & 0.75 \\
    IFT, Pseudoinverse \eqref{eq:gradient_approximation:pseudoinverse}              & 4.12 & 1.31 & 0.81 \\
    IFT, LM Constant \eqref{eq:gradient_approximation:lm}                           & 3.12 & 1.42 & 0.80 \\
    IFT, LM SVT \eqref{eq:gradient_approximation:lm}                                & 2.22 & 1.49 & 0.80 \\
    IFT, Residual Damping \eqref{eq:gradient_approximation:residual_damping}        & 1.58 & 1.76 & 0.81 \\
    IFT, Anisotropic Damping \eqref{eq:gradient_approximation:anisotropic_damping}  & 1.72 & 2.52 & 0.81 \\
    IFT, Full Newton \eqref{eq:gradient_approximation:full_newton}                  & 1.71 & 1.43 & 0.82 \\
    \midrule
    \multicolumn{4}{@{}l}{\textit{Proposed: IFT, Boundary Reachability}} \\
    IFT, Zero Gradients \eqref{eq:gradient_approximation:zero}                      & 1.64 & 1.20 & 0.75 \\
    IFT, Pseudoinverse \eqref{eq:gradient_approximation:pseudoinverse}              & 1.75 & 1.23 & 0.80 \\
    IFT, LM Constant \eqref{eq:gradient_approximation:lm}                           & 2.51 & 1.31 & 0.80 \\
    IFT, LM SVT \eqref{eq:gradient_approximation:lm}                                & 1.29 & 1.21 & 0.80 \\
    IFT, Residual Damping \eqref{eq:gradient_approximation:residual_damping}        & 1.15 & 1.24 & 0.81 \\
    IFT, Anisotropic Damping \eqref{eq:gradient_approximation:anisotropic_damping}  & 1.15 & 1.23 & 0.81 \\
    IFT, Full Newton \eqref{eq:gradient_approximation:full_newton}                  & 1.17 & 1.20 & 0.82 \\
    \bottomrule
    \end{tabular*}
\end{table}

The first experiment compares the new approach with an existing bespoke analytic IK solution that supports autodiff, to empirically demonstrate the correctness of our methodology and measure the performance cost of our more general approach.
The remaining experiments demonstrate the efficacy of our method together with the automatic IK solvers EAIK and IKFast.
This showcases the desired generalization: that our method can work with a much broader class of IK functions.

\subsection{Comparison with Forward-Mode Autodiff}
\label{sec:experiments:autodiff}

We evaluate the derivatives for the bimanual parameterization in isolation with low-level calculations, and we examine high-level performance on downstream tasks.
Let $\tilde q$ denote a point in the parameterized space and $q$ a point in C-space.

First, we examine speed and accuracy of the derivative calculations themselves.
We sample reachable configurations $\set{\tilde q_i}_{i=1}^{10\,000}$ (with rejection sampling).
For each configuration $q_i$, for a range of decision variable sizes $2^j$, we sample vectors of partial derivatives $\pd{\tilde q_i}{y_{ij}}\in\R^{2^j}$.
(Note that in these experiments, the partials do not correspond to any physical quantities; we are purely evaluating the speed and accuracy at the autodiff level.)
For each $(\tilde q_i,\pd{\tilde q_i}{y_{ij}})$, we compute $\pd{q_i}{y_{ij}}$ via forward-mode autodiff and our IFT technique.
The median error for each partial size is less than $10^{-13}$, and the 95th percentile is less than $10^{-12}$.
Autodiff is fastest up to partial size $2^6$, but for larger partial sizes, IFT is faster, due to constant cost of building the augmented Jacobian $D\FK_A(q)$.
Full numerical results are shown in \Cref{fig:autodiff_results}.

We also run downstream tasks that leverage these gradients.
For all of these experiments, we compare three reachability constraint formulations: direct reachability \eqref{eq:direct_reachability}, as in~\cite{cohn2024constrained}; probing functions, as in~\cite[\S{}IV]{cohn2026framework}; and boundary reachability \eqref{eq:boundary_reachability}, described in \Cref{sec:methodology:reachability}.
We compare autodiff to the IFT gradients, with each of the gradient regularization strategies presented in \Cref{sec:methodology:domain_extension}.
Note that we cannot use probing function reachability constraints with the IFT gradients, as they depend on implementation details unavailable to the IFT method.

First, we plan motions using graph of convex sets~\cite{marcucci2023motion}, which requires growing convex feasible subsets of the parameterized configuration space using IrisNp2, as presented in~\cite{werner2026faster}.
A key step of IrisNp2 is solving a \emph{counterexample search program} to find configurations for which a constraint (e.g. reachability) is violated.
Thus, applying IrisNp2 with an IK parameterization requires high-quality gradients outside the reachable workspace.
We also apply trajectory optimization~\cite[\S6.2]{russtedrake2024manipulation} (abbreviated ``Trajopt'') using bidirectional RRT~\cite{kuffner2000rrt} plus shortcutting~\cite{sekhavat1998multilevel} to obtain the initial guess.
An example trajectory is shown in \Cref{fig:iiwa_swept_volume}.
Finally, we retime trajectories to satisfy dynamics limits with TOPPRA~\cite{pham2018new}.

Runtimes are presented in \Cref{tab:downstream_timing}.
IrisNp2 runtimes are averaged over 17 seed points, Trajopt is averaged over 10 solves (each with a different RRT initial guess), and TOPPRA is averaged over the GCS, RRT plus shortcut, and Trajopt trajectories.
The Trajopt runtimes are comparable for most formulations; since it is given a feasible initial guess, its performance depends less on the gradient validity outside the reachable workspace.
TOPPRA does not use reachability constraints; the runtime increase is due to the higher cost of computing IFT gradients for small partial vectors (\Cref{fig:autodiff_results}).

As for IrisNp2, the direct reachability constraint leads to much higher runtime with the IFT gradients, due to their inaccuracy.
Our experimental results suggest that strategies which leverage the residual information (residual damping, anisotropic damping, and the full Newton method) are the most performant, with smallest difference in runtime performance against the bespoke baseline.
The boundary reachability constraint performs significantly better for IrisNp2, with the IFT runtime comparable to autodiff.

In summary, with appropriate gradient regularization, the more general IFT approach only suffers a minor runtime increase over bespoke autodiff gradient implementations.

\subsection{Using Generic IK Solvers}
\label{sec:experiments:generic}

We now demonstrate that our framework can be used to apply existing IK solvers without modification.

\subsubsection{Using a Solution Generated by EAIK}
\label{sec:experiments:generic:eaik}

\begin{figure}
    \centering
    \begin{subfigure}{\linewidth}
        \centering
        \includegraphics[width=\linewidth]{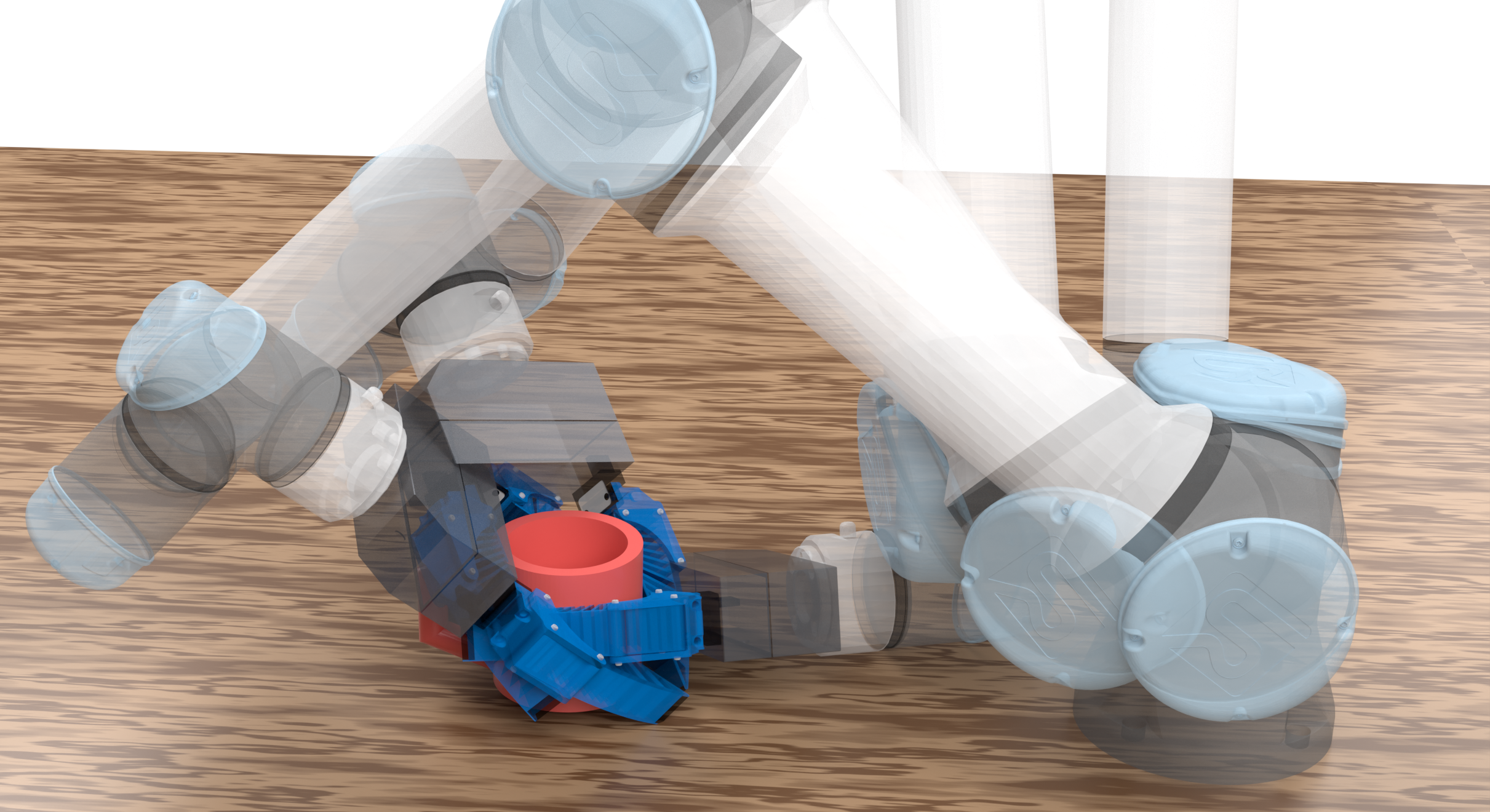}
    \end{subfigure}\break
    \begin{subfigure}{\linewidth}
        \centering
        \includegraphics[width=\linewidth]{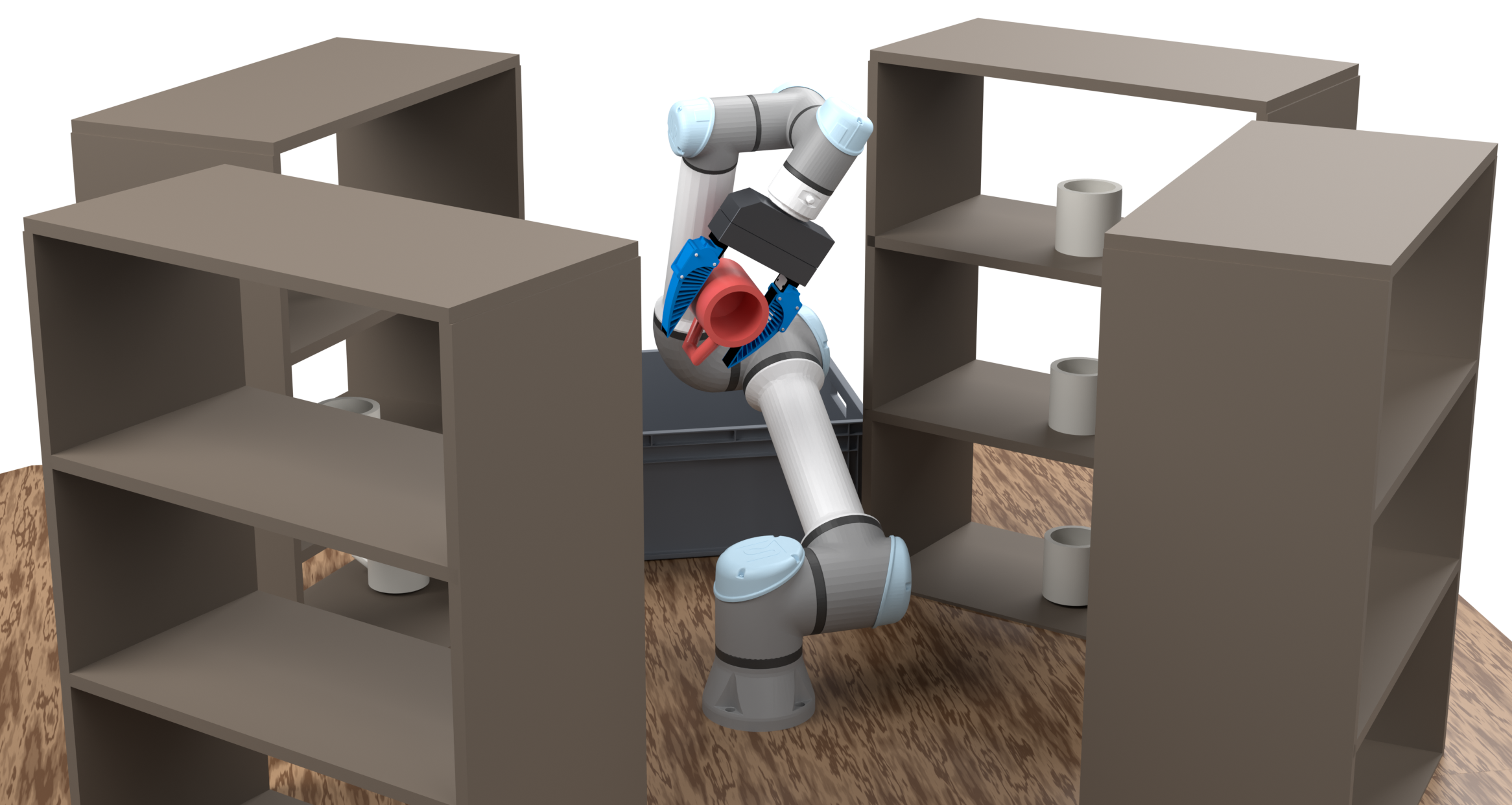}
    \end{subfigure}
    \caption{
        Above: three valid grasps for the EAIK grasp selection experiment setup, visualizing the variety of grasps encompassed by our optimization formulation.
        Below: the full setup, with added shelf obstacles.
    }
    \label{fig:grasp_selection_setup}
\end{figure}

Our first experiment using a generic solver without modification applies EAIK~\cite{ostermeier2025automatic}, which automatically derives IK solutions for a broad class of 6R robot arms.
EAIK uses a subproblem approach~\cite{elias2025ik}, each of which returns least-squares solutions when infeasible; thus, EAIK fits into our domain extension framework.
In practice, we found that EAIK solutions were sorted by branch, except when the least-squares solutions led to solution count changes.
Because the solver reports when a least-squares solution was used, tracking the branch index change could be handled in a straightforward manner.

We replicate an optimization-based grasp selection experiment~\cite[\S{}V.B]{cohn2026framework}, but using a UR5e.
The experiment setup (including examples of valid grasps and the obstacle layout) is shown in \Cref{fig:grasp_selection_setup}.
We transcribe the grasp selection problem as an optimization IK problem:
\begin{subequations}
\label{eq:grasp_selection}
\setlength{\jot}{1pt}
\begin{align}
        \min_{q\in\R^d} \colsep & \norm{q}_2^2
            \label{eq:grasp_selection:cost}\\[-3pt]
        \mathrm{s.t.} \colsep & p_{\lb} \le \monogram{p}{M}{G}(q) \le p_{\ub}
            \label{eq:grasp_selection:mug_constraint}\\
        \colsep & q_{\lb}\le q\le q_{\ub},
            \label{eq:grasp_selection:joint_limits}\\
        \colsep & q\ptxt{ is collision free},
            \label{eq:grasp_selection:min_distance}
\end{align}
\end{subequations}
where $M$ is the body frame of the manipuland and $\monogram{p}{M}{G}(q)$ denotes the position of the gripper in the manipuland frame.
We set the joint limits to $\pm\pi$.
\eqref{eq:grasp_selection:mug_constraint} enforces that a point between the fingers lie along the centerline of the mug, effectively a nonlinear equality constraint.
This can be rewritten as an optimization problem in minimal coordinates space via the IK change-of-variables:
\begin{subequations}
\label{eq:grasp_selection_parameterized}
\setlength{\jot}{1pt}
\begin{align}
    \min_{(\monogram{p}{M}{G},\monogram{o}{M}{G})\in\SE(3)} \colsep & \|\EIK(\monogram{p}{M}{G},\monogram{o}{M}{G})\|_2^2
            \label{eq:grasp_selection:parameterized:cost}\\[-3pt]
        \mathrm{s.t.} \colsep & p_{\lb} \le \monogram{p}{M}{G} \le p_{\ub}
            \label{eq:grasp_selection:parameterized:mug_constraint}\\
        \colsep & q_{\lb}\le \EIK(\monogram{p}{M}{G},\monogram{o}{M}{G})\le q_{\ub},
            \label{eq:grasp_selection:parameterized:joint_limits}\\
        \colsep & \EIK(\monogram{p}{M}{G},\monogram{o}{M}{G})\ptxt{ is collision free},
            \label{eq:grasp_selection:parameterized:min_distance}\\
        \colsep & (\monogram{p}{M}{G},\monogram{o}{M}{G})\ptxt{ is reachable},
            \label{eq:grasp_selection:parameterized:reachable}
\end{align}
\end{subequations}
where we have separated the pose of the gripper into the position $\monogram{p}{M}{G}$ and orientation $\monogram{o}{M}{G}$ for notational convenience.

We generate 100 problem instances with feasible goal states via rejection sampling in joint space.
(The mug is placed in the gripper, and the sample is accepted if the whole system is collision-free.)
For each pose, we solve the optimization problem from 10 randomly-sampled, collision-free initializations.
Initializations are kept consistent between the C-space and minimal formulation.
(This also determines the IK branch for the minimal formulation.)

We compare the ordinary configuration-space formulation \eqref{eq:grasp_selection} with the end-effector parameterized formulation \eqref{eq:grasp_selection_parameterized}, using both the direct reachability and boundary reachability constraints, with SNOPT~\cite{gill2005snopt} as the optimizer.
For direct reachability, we did not square the norm as in \eqref{eq:direct_reachability}; this decision empirically led to better performance on this problem instance.
We use IFT residual damping to compute $D\EIK$.

Results are in \Cref{tab:grasp_ik}.
The IFT formulations achieved higher success rates than the C-Space baseline, but found slightly higher-cost solutions, likely due to the distortion of the objective in the minimal coordinates.
The median runtimes were comparable, but the mean runtimes were worse for the IFT formulations, suggesting that it is generally easier for optimizers to certify convergence in the old formulation.

One possible explanation for the IFT formulation struggling to converge on certain instances is the approximate gradients outside the reachable workspace.
Line-search methods, used by both SNOPT and IPOPT, are known to struggle with convergence when gradients are inexact, so a trust region solver may be more performant in this context~\cite{carter1991global}.

\begin{table*}[!t]
    \centering
    \renewcommand{\arraystretch}{0.9}
    \caption{
        UR-5e grasp selection IK experimental results.
        The number in parentheses is restricted to three-way mutual successes.
    }
    \label{tab:grasp_ik}
    \begin{tabular}{lcccccc}
    \toprule
    \multicolumn{1}{c}{\multirow{2}{*}{\textbf{Formulation}}} & \multicolumn{2}{c}{\textbf{Success Rate (Multistart)}} & \multirow{2}{*}{\textbf{Mean Cost}} & \multirow{2}{*}{\textbf{Median Cost}} & \multirow{2}{*}{\textbf{Mean Runtime}} & \multirow{2}{*}{\textbf{Median Runtime}} \\
    \cmidrule(lr){2-3}
    & \makebox[2.2cm][c]{\textbf{1 init}} & \makebox[2.2cm][c]{\textbf{3 init}} & & & & \\[-0.5ex]
    \midrule
    \eqref{eq:grasp_selection}, C-Space Baseline            & 46.5\% & 86\% & 12.68 (12.81) & 10.94 (11.11) & 0.357 (0.365) & 0.105 (0.096) \\
    \eqref{eq:grasp_selection_parameterized}, IFT, Direct   & 61.4\% & 93\% & 13.30 (12.57) & 12.65 (11.95) & 0.719 (0.720) & 0.100 (0.083) \\
    \eqref{eq:grasp_selection_parameterized}, IFT, Boundary & 63.2\% & 90\% & 13.53 (13.03) & 12.83 (12.66) & 0.776 (0.825) & 0.110 (0.095) \\
    \bottomrule
    \end{tabular}
\end{table*}

\subsubsection{Using a Solution Generated by IKFast}
\label{sec:experiments:generic:ikfast}

In our second experiment, we use IKFast to derive a solution for the arms of a Rainbow RB-Y1 robot.
We then construct a parameterization that enables it to move objects with both hands, while maintaining a constant relative end-effector transform.
In particular, given analytic IK solutions $\IK_\LEFT$ and $\IK_\RIGHT$ for the left and right arms respectively, our parameterization is
\begin{equation*}
    \phi:(q_\base, q_\torso, \monogram{X_\LEFT}{W}{G}, \psi_\LEFT, \monogram{X_\RIGHT}{W}{G}, \psi_\RIGHT)\mapsto q_\full,
\end{equation*}
where $q_\base \in \R^3$ and $q_\torso \in \R^6$ are the joint angles of the RB-Y1's base and torso respectively, $\psi_\LEFT,\psi_\RIGHT\in\R$ are the self-motion parameters of the left and right arms, and $W$ is the world frame.
$q_\base$ is kept fixed in this experiment.

The robot's task is to pick up a box and place it on a nearby table from a set of starting positions arranged in a $4 \times 5$ grid (points spaced 3cm apart) on the floor in front of the robot's base, for a total of 20 trials.
Each pick-and-place trajectory is composed of six parts, planned sequentially:

\begin{enumerate}
    \item An \textit{unconstrained} motion plan from the robot's home position to a \textit{pre-grasp} pose where the grippers are placed 8cm above the box.
    \item An \textit{unconstrained} motion to the grasp pose.
    \item A \textit{constrained} motion plan to a nominal `lift' pose where the robot is standing straight up.
    \item A \textit{constrained} motion plan to the place location. A swept volume of this motion is shown in \Cref{fig:teaser}.
    \item An \textit{unconstrained} motion to a post-grasp pose where the grippers are 8cm above the box.
    \item An \textit{unconstrained} motion plan to the home position.
\end{enumerate}

We compute a set of grasp configurations using an optimization IK formulation~\cite[Appx.~C.5]{cohn2026framework}, and then the ``best'' grasps are selected using a cheap meta-heuristic and an expensive plannability check.
Motions are planned with a bidirectional RRT (BiRRT) with shortcutting (unconstrained motions operate in C-space, constrained motions use the parameterization framework).
The pre-grasp and post-place motions attempt a straight line in C-space, falling back to BiRRT if it fails.
These trajectories are then used as initial guesses for Trajopt and then retimed with TOPPRA.
For all these processes, we enforce static stability by constraining the center of mass of the robot to stay over the support polytope~\cite[\S{}48.3.1]{siciliano2008springer}.
All calculations are given gradients derived using the presented IFT method.
To approximate the analytic IK domain extension, we use a bisection search between the unreachable configuration and a nominal end-effector position, together with residual damping \eqref{eq:gradient_approximation:residual_damping}.

\begin{table}[!t]
    \centering
    \renewcommand{\arraystretch}{0.9}
    \caption{
        RB-Y1 box pickup experimental results.
    }
    \label{tab:full_body_ik}
    \begin{tabular}{lcc}
        \toprule
        \textbf{Metric} & \textbf{Mean} & \textbf{Max} \\
        \midrule
        \textit{Planning Statistics} & & \\
        Total Runtime (s) & 56.1 & 135.0 \\
        Optimization IK Runtime (s) & 19.4 & 84.0 \\
        Trajectory Optimization Runtime (s) & 5.6 & 29.6 \\
        \midrule
        \textit{Hardware Statistics} & & \\
        Trajectory Duration (s) & 54.1 & 63.6 \\
        Measured Constraint Violation (mm) & 0.555 & 2.235 \\
        Measured Constraint Violation (mrad) & 1.597 & 6.809 \\
        \bottomrule
    \end{tabular}
\end{table}

Results are reported in \Cref{tab:full_body_ik}, and videos of all hardware rollouts are available at the \href{https://cohnt.github.io/inverse-function-theorem-parameterization/}{project website}.
The robot can consistently compute a path that picks up the box from various configurations and places it on the table.
The planned trajectory satisfies the kinematic constraint up to floating point tolerance.
Thus, by just tracking the plan with joint position control, the robot is able to successfully deliver the object to the table without damage or dropping.

\section{Discussion}
\label{sec:discussion}

We have presented a novel and significantly more general approach for differentiating through IK mappings, based on the inverse function theorem.
Our method avoids the onerous and potentially-prohibitive requirement to modify analytic IK functions to support autodiff, lowering the barrier to entry of constrained trajectory optimization in minimal coordinates.
We further introduce optimization-amenable domain extensions and reachability constraints for analytic IK functions, achieving close performance to bespoke IK solutions.
In many robotics contexts, we want to reason about the robot as a floating hand -- our approach yields the necessary gradients to solve optimization problems with that perspective.
And while DiffIK only reasons locally about redundancy resolution, potentially getting stuck while tracking challenging end-effector trajectories, our global approach produces motions that avoid joint limits, workspace boundaries, and kinematic singularities.

Furthermore, our results inform high-level design choices made when solving robotic optimization problems with IK parameterizations.
Any well-defined IK function which guarantees continuity of solutions can be deployed, as the gradient calculations are now implementation-independent.
Residual-based regularization strategies only require a small amount of tuning for performance, and boundary reachability can robustly handle non-reachable optimizer iterates.

Open questions remain in this space, especially related to IK solution tracking for solvers that do not globally sort IK solutions into ``branches'' -- examples include IK-Geo's subproblems 5 and 6~\cite{elias2025ik} and polynomial/eigenvalue IK methods~\cite{husty2007new}.
For cuspidal manipulators~\cite{elias2025path}, such challenges are present independent of the IK algorithm, and path planning for such robots in the minimal-coordinates framework remains an open problem.

\section*{Acknowledgement}

We thank Aditya Agarwal and Lucy Cai for their valuable contributions to the RB-Y1 communication infrastructure, and Shrutheesh Iyer for his insights on kinematics.

\bibliographystyle{IEEEtran}
\bibliography{ref.bib}

\end{document}